\documentclass[runningheads]{llncs}

\usepackage{eccv}

\usepackage{eccvabbrv}

\usepackage{graphicx}
\usepackage{booktabs}

\usepackage[accsupp]{axessibility}  

\usepackage[pagebackref,breaklinks,colorlinks,citecolor=eccvblue]{hyperref}

\usepackage{orcidlink}

\usepackage{bm}
\usepackage{multirow}
\usepackage{array}

\usepackage{hyperref}
\usepackage{xcolor}
\usepackage{booktabs}
\hypersetup{
    colorlinks=true,
    citecolor=green,
    linkcolor=red,
    urlcolor=green,
}

\newcolumntype{C}[1]{>{\centering\arraybackslash}p{#1}}
\newcolumntype{R}[1]{>{\raggedleft\arraybackslash}p{#1}}
\newcolumntype{L}[1]{>{\raggedright\arraybackslash}p{#1}}

\begin{document}

\title{MaskDiffusion: Exploiting Pre-trained Diffusion Models for Semantic Segmentation} 

\titlerunning{Abbreviated paper title}

\author{Yasufumi Kawano\inst{1}\orcidlink{0000-0002-2252-1910} \and
Yoshimitsu Aoki\inst{1}\orcidlink{0000-0001-7361-0027}}

\authorrunning{Y.~Kawano and Y.~Aoki.}

\institute{Keio University, Japan \\
\email{ykawano@aoki-medialab.jp, aoki@elec.keio.ac.jp}}

\maketitle

\begin{abstract}
Semantic segmentation is essential in computer vision for various applications, yet traditional approaches face significant challenges, including the high cost of annotation and extensive training for supervised learning. Additionally, due to the limited predefined categories in supervised learning, models typically struggle with infrequent classes and are unable to predict novel classes. To address these limitations, we propose MaskDiffusion, an innovative approach that leverages pretrained frozen Stable Diffusion to achieve open-vocabulary semantic segmentation without the need for additional training or annotation, leading to improved performance compared to similar methods. We also demonstrate the superior performance of MaskDiffusion in handling open vocabularies, including fine-grained and proper noun-based categories, thus expanding the scope of segmentation applications. Overall, our MaskDiffusion shows significant qualitative and quantitative improvements in contrast to other comparable unsupervised segmentation methods, i.e. on the Potsdam dataset (+10.5 mIoU compared to GEM) and COCO-Stuff (+14.8 mIoU compared to DiffSeg). All code and data will be released at \url{https://github.com/Valkyrja3607/MaskDiffusion}.
\end{abstract}

\section{Introduction}
\label{sec:intro}

\begin{figure}[t]
    \centering
    \includegraphics[width=0.8\linewidth]{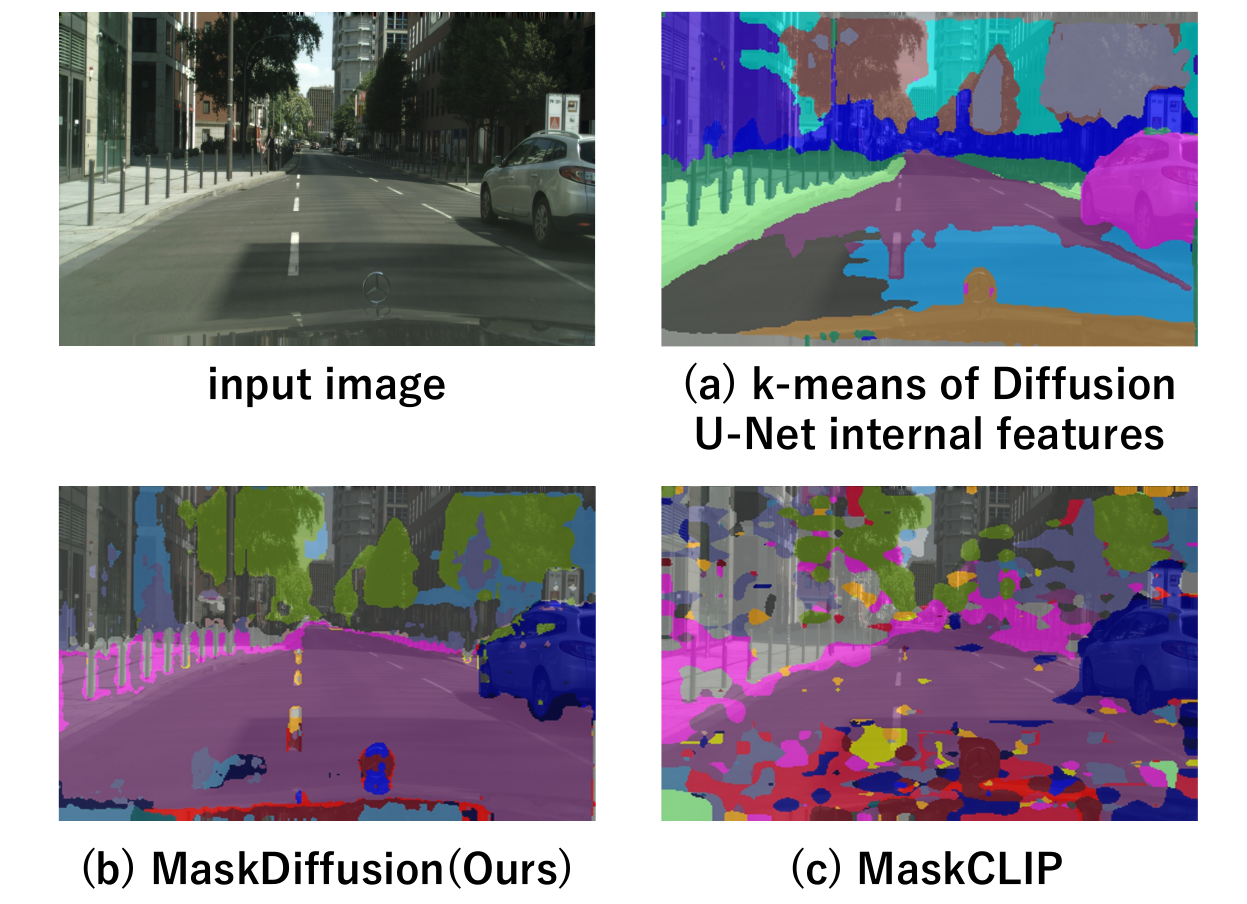}
    \caption{\textbf{Comparison of MaskDiffusion with previous method MaskCLIP~\cite{maskclip} on a Cityscapes~\cite{cityscapes} image.} k-means clustering on the internal features of the Diffusion U-Net (a) shows that each determined cluster roughly partitions the image according to some classes, indicating that the semantic information is well preserved. MaskDiffusion (b) yields well-partitioned segments consistent with the shape of the object and exhibits minimal noise. In comparison, MaskCLIP~\cite{maskclip} (c) results in smaller and noisy segments.}
    \label{fig:compare}
\end{figure}

Semantic segmentation, a fundamental task in computer vision, assigns class labels to every pixel in an image. Its applications span across diverse domains, including automated driving and medical image analysis. 
Despite its significance, current semantic segmentation methods still face several critical challenges.
First of all, these methods tend to be costly, requiring pixel-level annotation as well as extensive training. Secondly, since supervised learning relies on a pre-defined set of categories, detecting extremely rare or even completely new classes during prediction becomes virtually impossible.

In this paper, we address these limitations by combining two related tasks, namely unsupervised and open-vocabulary semantic segmentation. Unsupervised semantic segmentation~\cite{picie,stego,hp} avoids costly annotation by leveraging representations obtained through a model ~\cite{dino,dinov2} which has been trained on a different task. On the other hand, open-vocabulary semantic segmentation~\cite{sam, openseg, odise, maskclip, openseg2} allows recognizing a wide array of categories through natural language and is not bound to a pre-defined set of categories. To integrate these two approaches, we leverage the internal feature space as well as the attention maps of a pre-trained Stable Diffusion model.

Diffusion models~\cite{sd, ddpm, imagen} trained on large-scale datasets have revolutionized the field of image generation, which can be largely attributed to conditioning with text embeddings derived from large-scale pre-trained visual-language models such as CLIP~\cite{clip}. We hypothesize that these diffusion models have mastered a wide variety of open vocabulary concepts that could be used for dense prediction tasks, particularly for semantic segmentation. In previous works~\cite{odise, openseg}, the internal features of a frozen diffusion model have already been used for semantic segmentation after some additional training. 
Based on these observations, we take a closer look at the internal features of diffusion models for semantic segmentation and introduce \textbf{MaskDiffusion}, which achieves effective segmentation in the wild \textbf{without} additional training.

The strengths of our MaskDiffusion are manifold.
First, it eliminates the need for pixel-by-pixel annotations typically required by popular semantic segmentation techniques. Second, it has the ability to segment any class of objects, distinguishing it from traditional diffusion-based methods~\cite{odise,ddpmseg,segdiff}. 

As shown in Figure~\ref{fig:compare}, applying k-means clustering on the internal features of the U-Net in Stable Difusion already provides a rough yet consistent segmentation, whereas MaskCLIP~\cite{maskclip} fails to uniformly segment the image. We are able to further improve segmentation results leveraging internal features with our proposed MaskDiffusion.

In addition, based on the observation that internal features are useful for segmentation, we propose an unsupervised segmentation method, called \textbf{Unsupervised MaskDiffusion}.
Unsupervised MaskDiffusion, unlike MaskDiffusion, operates solely on image inputs without requiring class candidate prompts. This is why it is "Unsupervised".
Unsupervised MaskDiffusion can segment objects of the same class by spectral clustering~\cite{spectral}, computing a Laplacian matrix from the similarity between pixels of internal features.

Our contributions are the following:
\begin{enumerate}
    \item We analyze the internal features of diffusion models and show that they are useful for semantic segmentation. The results of k-means classification of internal features are comparable to those of unsupervised mIoU, a conventional unsupervised segmentation method.
    \item We introduce MaskDiffusion and Unsupervised MaskDiffusion achieving compelling segmentation results for all categories in the wild without any additional training.
    \item MaskDiffusion model outperforms GEM~\cite{gem} by 10.5 mIoU on the Potsdam dataset~\cite{potsdam}, demonstrating the superior segmentation performance of our proposed approach. Moreover, the Unsupervised MaskDiffusion model surpasses the performance of DiffSeg~\cite{diffseg} by 14.8 mIoU on the COCO-Stuff dataset~\cite{coco}, as measured by unsupervised mIoU.
    
\end{enumerate}

\section{Related Work}
\subsection{Diffusion Model}
Diffusion models, particularly Denoising Diffusion Probabilistic Models (DDPM)~\cite{ddpm}, excel in generating high-quality images by iteratively adding and then estimating and removing Gaussian noise from initial images. Furthermore, the implementation of Stable Diffusion utilizes a Latent Diffusion Model (LDM)~\cite{sd}, which combines Variational Autoencoder (VAE)~\cite{vae}, UNet~\cite{unet}, and CLIP~\cite{clip}, a powerful visual language model. Stable Diffusion integrates CLIP, a text encoder, to represent the text content in the image. This integration involves conditioning the latent space in the UNet~\cite{unet} architecture with text, effectively enabling the representation of textual content in synthesized images. 
Notably, UNet's training on the LAION-5B dataset~\cite{laion}, with its 5 billion image-text pairs, highlights the importance of large-scale data for robust image-text generation.

\begin{table}[t]
  \caption{\textbf{Relationship with related works.} }
  \label{tab:relationship}
  \centering
  \scalebox{0.78}[0.78]{
  \begin{tabular}{l | c | c | c | c | c}
  \hline
    \multirow{2}{*}{Method} & Backbone & Additional & Language & Class & Mapping \\
     &  Pretraining & Training & Dependency & Identifiability & Consistency \\
     \hline
     \multicolumn{6}{l}{Open Vocabulary Segmentation} \\
     \hline
     ODISE~\cite{odise} & SD~\cite{sd} internal feature & $\surd$ & $\surd$ & $\surd$ & $\surd$ \\
     MaskCLIP~\cite{maskclip} & CLIP~\cite{clip} & - & $\surd$ & $\surd$ & $\surd$ \\
     MaskDiffusion (Ours)& SD~\cite{sd} internal feature & - & $\surd$ & $\surd$ & $\surd$ \\
     \hline
     \multicolumn{5}{l}{Unsupervised Segmentation} \\
     \hline
     STEGO~\cite{stego} & DINO~\cite{dino} & $\surd$ & - & - & $\surd$ \\
     DiffSeg~\cite{diffseg} & SD~\cite{sd} self-attention & - & - & -  & - \\
     Unsupervised MaskDiffusion (Ours)& SD~\cite{sd} internal feature & - & - & - & - \\
    \hline
  \end{tabular}
  }
\end{table}

\subsection{Semantic Segmentation}

Semantic segmentation involves pixel-wise class labeling, commonly using convolutional neural networks~\cite{deeplab, long} or vision transformers~\cite{transformer} for end-to-end training. These methods, while effective, depend on extensive labeled data and significant computational resources, and are limited to predefined categories. Thus, unsupervised, and domain-flexible approaches have recently gained importance.

Unsupervised semantic segmentation~\cite{iic,picie,stego,hp} attempts to solve semantic segmentation without using any kind of supervision.
STEGO~\cite{stego} and HP~\cite{hp} optimize the head of a segmentation model using image features obtained from a backbone pre-trained by DINO~\cite{dino}, an unsupervised method for many tasks.
However, unsupervised semantic segmentation clusters images by class but cannot identify each cluster's class.
In contrast, our MaskDiffusion distinguishes classes without extra annotation or training. 

Open vocabulary semantic segmentation, crucial for segmenting objects across domains without being limited to predefined categories, has seen notable advancements with the introduction of key methodologies~\cite{sam,maskclip,gem,clipseg,reco, zeroshot,seem}.
MaskCLIP~\cite{maskclip}, GEM~\cite{gem} and CLIPSeg~\cite{clipseg}, based on the CLIP~\cite{clip} visual language model, have advanced open vocabulary semantic segmentation. In particular, MaskCLIP~\cite{maskclip} and GEM~\cite{gem} segment various categories without additional training. These two approaches, reducing annotation and training costs and overcoming category limitations, have inspired a new direction in our research, motivating us to explore similar problem formulations. Furthermore, ODISE~\cite{odise} has extended the capabilities of pre-trained diffusion models, incorporating additional training of a segmentation network to achieve open-vocabulary panoptic segmentation. The application of image generation models to segmentation is discussed in detail in Sec~\ref{sec:genemodelseg}.

Table~\ref{tab:relationship} shows the pre-trained model used (Backbone Pretraining), whether additional training is used (Additional Training), whether text input (prompt) is required for segmentation (Language Dependency), whether the same class can be assigned to the same segment index across images (Class Identifiability), and whether there is consistency in the index mapped across images (Mapping Consistency).
In particular, for Class Identifiability and Mapping Consistency, unsupervised segmentation methods do not explicitly know the mapping between indices and classes in the model's output images. STEGO~\cite{stego}, however, have a consistent mapping across images. In contrast, DiffSeg~\cite{diffseg} and our Unsupervised DiffusionMask output different indices for segments of the same class across images.

\subsection{Generative Models for Segmentation}
\label{sec:genemodelseg}
The use of image generative models, including Generative Adversarial Networks (GAN)~\cite{networkfree,gan1,gan2,gan3,gan4,gan5} and diffusion models~\cite{odise,ddpmseg,diff1,diff2,diff3,diff4,diffumask,segdiff,diffseg}, has been a focal point in various prior studies concerning semantic segmentation. 
DDPMSeg~\cite{ddpmseg} revealed that a generative model's internal representation correlates with visual semantics, aiding semantic segmentation but limited to a predefined set of labels. Addressing this, ODISE~\cite{odise} enables panoptic segmentation with open vocabularies. 
DiffSeg~\cite{diffseg}, similarly to our method, provides unsupervised segmentation without additional training by using KL divergence in UNet's self-attention maps for segmentation, but it cannot consistently map the same class to the same index across images.

Several methodologies~\cite{odise,ddpmseg,diff1,diff2,diff3,diff4} have leveraged diffusion models as a foundational backbone for segmentation tasks, often incorporating additional models and training to facilitate the segmentation process.

In alignment with the existing research landscape, we propose an open-vocabulary semantic segmentation method without additional training.

\begin{figure*}[t]
    \centering
    \begin{minipage}[b]{\linewidth}
    \includegraphics[width=\linewidth]{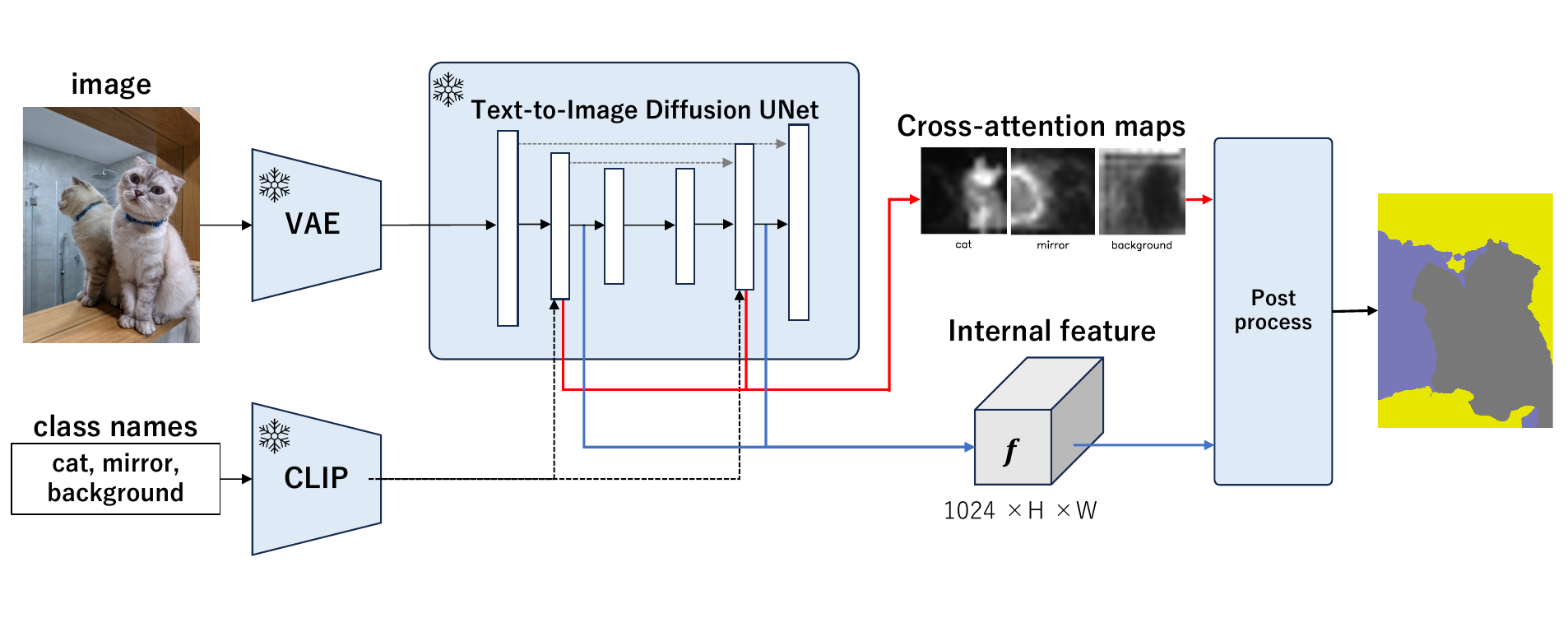}
    \vspace{-1mm}
    \caption{\textbf{High-level overview of our MaskDiffusion architecture.} MaskDiffusion uses a frozen pre-trained diffusion model. The UNet is given latent images compressed by a VAE as well as text prompts embedded by CLIP. The prompts are the names of all the potential classes to be segmented. The output of each layer of the U-Net is extracted as a concatenated internal feature $\mathbf{f}$ and a cross-attention map, which are subsequently post-processed into a segmentation image.}
    \label{fig:overview}
    \end{minipage}
\end{figure*}

\begin{figure}[t]
    \centering
    \includegraphics[width=0.85\linewidth]{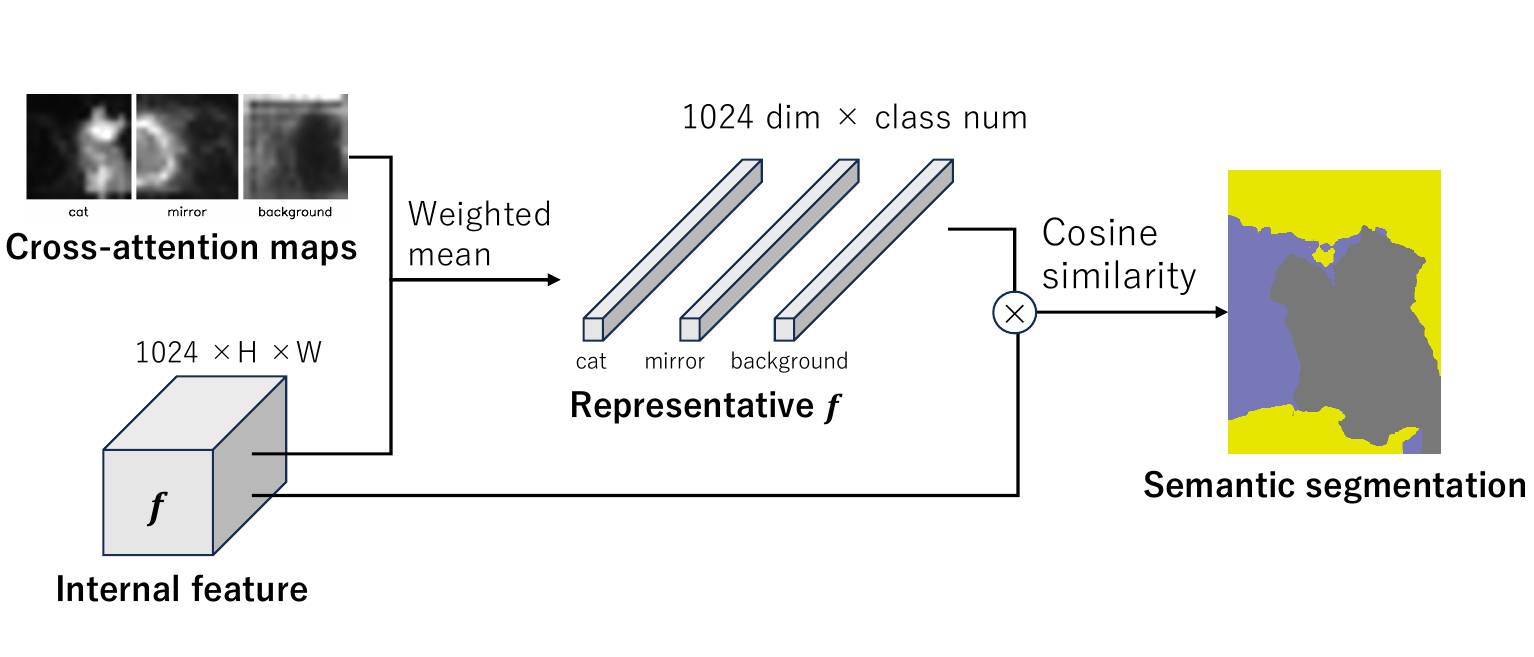}
    \vspace{-3mm}
    \caption{\textbf{Overview of the post processing step.} First, a representative $\mathbf{f}$ is computed for each category through a weighted average of $\mathbf{f}$ based on the values of the cross-attention map. In the next step, we determine the semantic segmentation result by evaluating the cosine similarity between $\mathbf{f}$ and the representative $\mathbf{f}$ for each category and then assign each $\mathbf{f}$ to the category that has the closest similarity.}
    \label{fig:postprocess}
\end{figure}

\section{Method}
Our research attempts to explore the applicability of diffusion models to semantic segmentation. We begin with a brief introduction to the architecture of diffusion models in Sec~\ref{sec:diffusion}, followed by a detailed description of the proposed MaskDiffusion in Sec~\ref{sec:maskdiff}.

\subsection{Diffusion Model Architecture}
\label{sec:diffusion}

Stable Diffusion~\cite{sd} follows  the two processes in diffusion models, namely the diffusion process and the inverse diffusion process. The former involves the introduction of Gaussian noise to a clean image, while the latter employs a single UNet architecture to eliminate the noise and restore the original image. The diffusion and inverse diffusion processes each consist of 1000 steps, and a single UNet is responsible for all time steps. During training, we estimate the noise as

\begin{align}
\label{lossdiff}
L_{LDM} := \mathbb{E}_{\varepsilon(x), y, \epsilon \sim \mathcal{N}(0,1),t} \left[ \| \epsilon - \epsilon_{\theta}(z_t, t, \tau_\theta(y)) \|^2_2 \right] ,
\end{align}

\noindent
where $x$ is the input image, $z$ and $\varepsilon(x)$ are the latent images, $\epsilon$ is the noise sampled from a normal distribution $\mathcal{N}(0,1)$, $t$ is the time step, $y$ is the text prompt, $\tau_\theta$ is a pre-trained text encoder and $\epsilon_{\theta}(z_t, t, \tau_\theta(y))$ is the predicted noise, which is the output of the UNet.

A significant aspect of Stable Diffusion~\cite{sd} lies in the utilization of a trained CLIP~\cite{clip} as the text encoder, facilitating the conversion of textual inputs into corresponding vectors. Embedding these vectors into the UNet architecture enables the integration of text-based conditioning during the denoising process, achieved through cross-attention mechanisms. Attention consists of three vectors: query ($Q$), key ($K$), and value ($V$), and is defined as 

\begin{align}
\label{qkv}
Q = W^{(i)}_Q \cdot \phi_i(z_t), K &= W^{(i)}_K \cdot \tau_{\theta}(y), V = W^{(i)}_V \cdot \tau_{\theta}(y), \nonumber\\
\text{Attention}(Q, K, V) &= \text{softmax}\left(\frac{QK^T}{\sqrt{d}}\right) \cdot V ,
\end{align}

\noindent
where d is the scaling factor, $\tau_{\theta}$ is the text encoder, $y$ is the text prompt, $N$ is the number of tokens in the input sequence, $\phi_i(z_t) \in \mathbb{R}^{N \times d_i}$ is a (flattened) intermediate
representation of the UNet implementing $\epsilon_{\theta}$ and $W^{(i)}_V \in \mathbb{R}^{d \times d_i}$, $W^{(i)}_Q \in \mathbb{R}^{d \times d_\tau}$, $W^{(i)}_K \in \mathbb{R}^{d \times d_\tau}$ are projection matrices.
The query is a vector created from the input image, while the key and value are created from the text vector.

\subsection{MaskDiffusion}
\label{sec:maskdiff}
Figure~\ref{fig:overview} shows our proposed method which we call MaskDiffusion, a novel approach that leverages a pre-trained diffusion model as a semantic segmentation model, eliminating the need for additional training. MaskDiffusion capitalizes on the observation that the intrinsic features derived from the internal layer of the UNet in Stable Diffusion~\cite{sd} inherently contain semantic information, as demonstrated in the experimental results outlined in Section~\ref{sec:internal}.
For each pixel of the input image we obtain an internal feature, denoted as $\mathbf{f} \in \mathbb{R}^{1024}$, by upsampling and combining the output of the internal layers as follows,

\begin{align}
\label{eq:f}
\mathbf{f}_i = \text{UNetLayer}_i(x_i, \tau_{\theta}(s)), \nonumber \\
\mathbf{f} = \text{Concatenate}(\mathbf{f}_1, ..., \mathbf{f}_n) ,
\end{align}

\noindent
where $x_{i}$ denotes the input to the internal UNet layer, $s$ represents the prompt and $n$ is the number of internal UNet layers. Note that $x_0$ is the input image and $x_1$ and beyond are the output of previous layers.

Due to its high dimensionality, determining the class of each pixel solely based on $\mathbf{f}$ proves to be a challenging task. Therefore, we propose to incorporate the information contained in the cross-attention maps in the UNet architecture into $\mathbf{f}$ in order to assign a class to each pixel.

\begin{equation}
\label{cross-attention}
\text{Cross-attention map}(Q, K) = \frac{QK^T}{\sqrt{d}}
\end{equation}

We define the cross-attention map as shown in Equation~\ref{cross-attention} based on the definition of the cross-attention in Equation~\ref{qkv}. Note that we do not use the softmax function in Equation~\ref{cross-attention} as it introduces undesirable normalization across classes, which disrupts the intrinsic relationship between pixels and consequently leads to poorer performance. 

Cross-attention establishes a relationship between the textual input prompts, which include the category names for segmentation, and the pixel-wise features of the image. In essence, the cross-attention map can be viewed as a preliminary semantic segmentation map, highlighting areas of potential object localization. However, using the cross-attention map as a standalone solution does not lead to accurate results as shown through our experiments, which are summarized in Table~\ref{tab:main}. Instead, we leverage the cross-attention map in combination with the extracted high-dimensional internal features, $\mathbf{f}$, to assign each pixel a corresponding class. 

Next, we explain how to incorporate the information of the cross-attention map into the internal features $\mathbf{f} \in \mathbb{R}^{1024 \times H \times W}$ as illustrated in Figure~\ref{fig:postprocess}. First, we compute a representative internal feature $\bar{\mathbf{f}}_c \in \mathbb{R}^{1024}$ for each category $c$ through a weighted average based on the values $a_{cnm} \in \mathbb{R}$ of the cross-attention map, as follows:

\begin{align}
\label{eq:pp}
\bar{\mathbf{f}}_{c} &= \frac{1}{A_c} \sum_{n,m} a_{cnm} \cdot \mathbf{f}_{nm}, \quad A_c=\sum_{n,m} a_{cnm}
\end{align}

In the subsequent step, the semantic map $\mathbf{s} \in \mathbb{R}^{H \times W}$ is obtained by calculating the cosine similarity for each pixel position ($n$, $m$) between the internal feature vector and the representative feature $\bar{\mathbf{f}}_c$ for each class and subsequently assigning the class with the highest similarity. Mathematically, this process is described as 

\begin{align}
\label{eq:pp2}
s_{nm} &= \underset{c}{\text{argmax}} \frac{\mathbf{f}_{nm}^T \cdot \bar{\mathbf{f}}_c}{\| \mathbf{f}_{nm} \| \cdot  \|\bar{\mathbf{f}_c} \|} ,
\end{align}

\noindent
and can be interpreted as assigning the category, represented by $\bar{\mathbf{f}}_c$, which exhibits the closest resemblance with the internal feature $\mathbf{f}_{nm}$ of the corresponding pixel.

\begin{figure}[t]
    \centering
    \begin{minipage}[b]{\linewidth}
    \includegraphics[width=\linewidth]{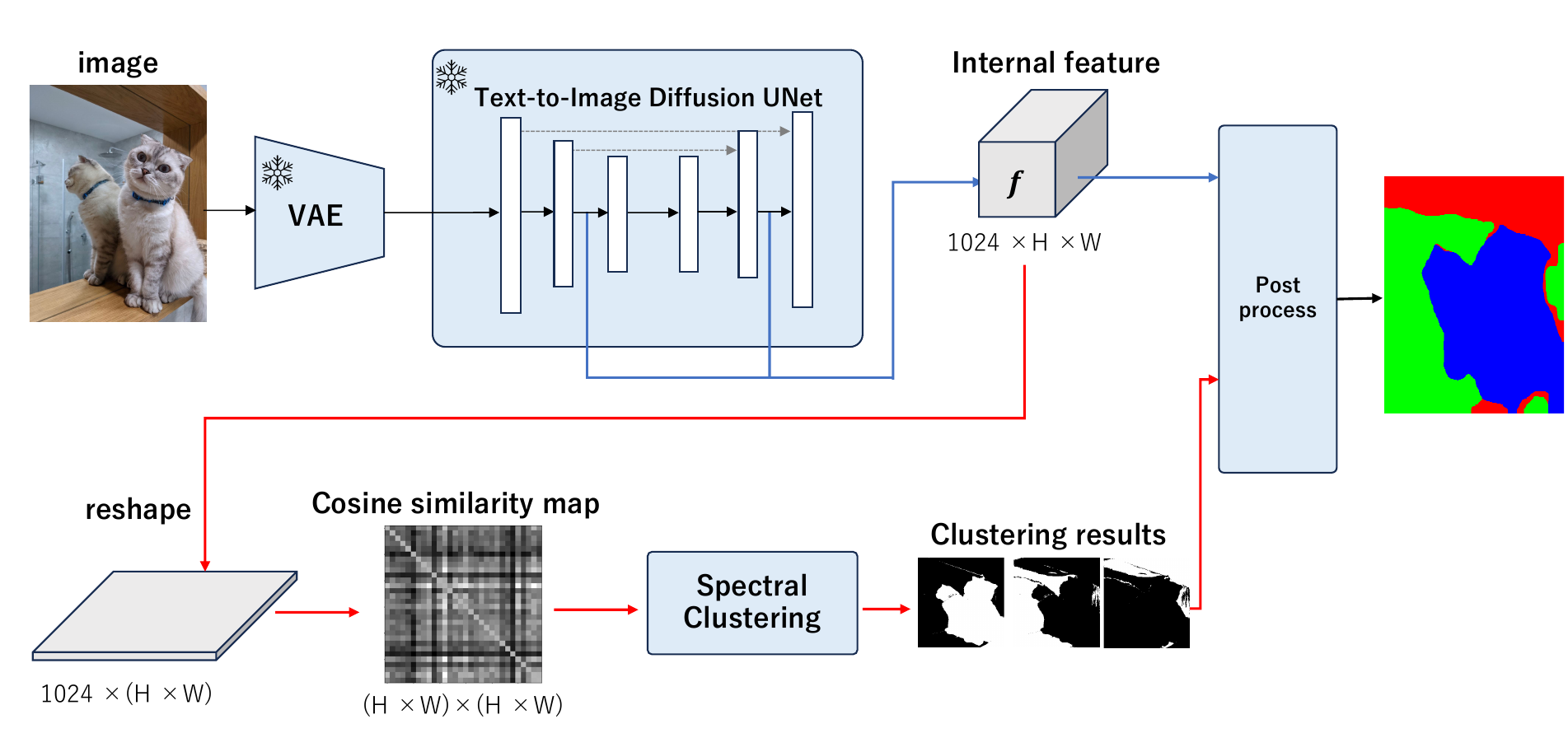}
    \caption{\textbf{Overview of Unsupervised MaskDiffusion architecture.} we employ spectral clustering~\cite{spectral} to include the spatial relationships of the internal features in the segmentation process.}
    \label{fig:umaskdiffusion}
    \end{minipage}
\end{figure}
\subsection{Unsupervised MaskDiffusion}
\label{sec:umaskdiff}
In our experimental results with MaskDiffusion, as detailed in Section~\ref{sec:main}, we found that the internal features seem to be quite powerful, while the cross-attention map demonstrates some limitations. As a consequence, we also explore an alternative approach that only uses internal features in the context of unsupervised segmentation. As such, this method does not require any prompts and only relies on images as input, which corresponds to the same setting as DiffSeg~\cite{diffseg}.

Unsupervised MaskDiffusion overview is shown in Figure~\ref{fig:umaskdiffusion}.
In order to cluster the obtained internal features in the text-free setting, we explore two different clustering methods, namely k-means and spectral clustering~\cite{spectral}. In particular, we employ spectral clustering~\cite{spectral} to include the spatial relationships of the internal features in the segmentation process. 
Based on the observation that internal features representing the same class should be close with respect to the distance in pixel space, we compute the similarity map $\mathbf{h} \in \mathbb{R}^{HW \times HW}$ between the reshaped internal features $\hat{\mathbf{f}} \in \mathbb{R}^{HW \times 1024}$ as

\begin{align}
\label{sim_map}
\mathbf{h}_{ij} = \frac{\mathbf{\hat{\mathbf{f}}_{i}}^T \cdot \mathbf{\hat{\mathbf{f}}_{j}}}{\| \mathbf{\hat{\mathbf{f}}_{i}} \| \cdot  \|\mathbf{\hat{\mathbf{f}}_{j}} \|} ,
\end{align}

\noindent
where $i$ and $j$ denote the pixel location. A large value of $\mathbf{h}_{ij}$ in this map indicates that the pixels at $i$ and $j$ very likely belong to the same class.
Note that we set $H$ and $W$ to 100 and accordingly resize $\mathbf{f}$ to $100 \times 100$ to reduce computational complexity.

Next, we derive a Laplacian matrix from the similarity map and classify clusters based on their smallest eigenvectors. The clustering results are then post-processed using $\mathbf{f}$, where the clustering results replace the cross-attention map in Equation~\ref{eq:pp} to produce the final segmentation image.
Note that the values of the spectral clustering~\cite{spectral} are either 0 or 1, whereas the cross-attention map consists of values between 0 and 1.

\section{Experiment}
First, we present the implementation details and then compare our results to previous methods in Section~\ref{sec:main} and \ref{sec:internal}.
Next, we evaluate the open vocabulary aspect in Section~\ref{sec:web}. Finally, we justify the construction of MaskDiffusion through an ablation study in Section~\ref{sec:ablation}.

\subsection{Implementation Details}
\label{sec:setting}
\noindent
For our implementation of MaskDiffusion, we employed a frozen Stable Diffusion model~\cite{sd} that was pre-trained on a subset of the LAION~\cite{laion} dataset, utilizing CLIP~\cite{clip}'s ViT-L/14 architecture to condition the diffusion process on text input.
We extracted internal features and cross-attention maps from Stable Diffusion's three UNet blocks and combined them by resizing and concatenating. In addition, images exceeding $512 \times 512$ are segmented into $512 \times 512$ patches prior to model input.
Furthermore, we set the time step for the diffusion process to $t = 1$, representing the last denoising step, as this ensures an efficient processing time of less than 2 seconds per image.
Our model works with a GPU memory of 15 GB.

\begin{table*}[t]
  \caption{\textbf{Comparison of MaskCLIP~\cite{maskclip}, GEM~\cite{gem}, and our MaskDiffusion.} We evaluate on Potsdam~\cite{potsdam}, Cityscapes~\cite{cityscapes}, PascalVOC~\cite{voc} and COCO-Stuff~\cite{coco} and report the mIoU.}
  \label{tab:main}
  \centering
  \scalebox{1.0}[1.0]{
  \begin{tabular}{l | c | c | c | c}
  \hline
    \multirow{2}{*}{Method} & Potsdam~\cite{potsdam} & Cityscapes~\cite{cityscapes}& PascalVOC~\cite{voc} & COCO-Stuff~\cite{coco} \\
    \cline{2-5}
     & 6classes & 19class & 20class & 171class \\
     \hline
     MaskCLIP~\cite{maskclip}  & 10.4 & 15.5 & 28.6 & 1.6 \\
     GEM~\cite{gem} & 10.7 & 16.3 & 26.5 & 0.8 \\
     MaskDiffusion (Ours) & \textbf{21.2} & \textbf{17.1} & \textbf{29.9} & \textbf{5.5} \\
    \hline
  \end{tabular}
  }
\end{table*}
\begin{figure*}[t]
    \begin{minipage}[b]{\linewidth}
    \includegraphics[width=\linewidth]{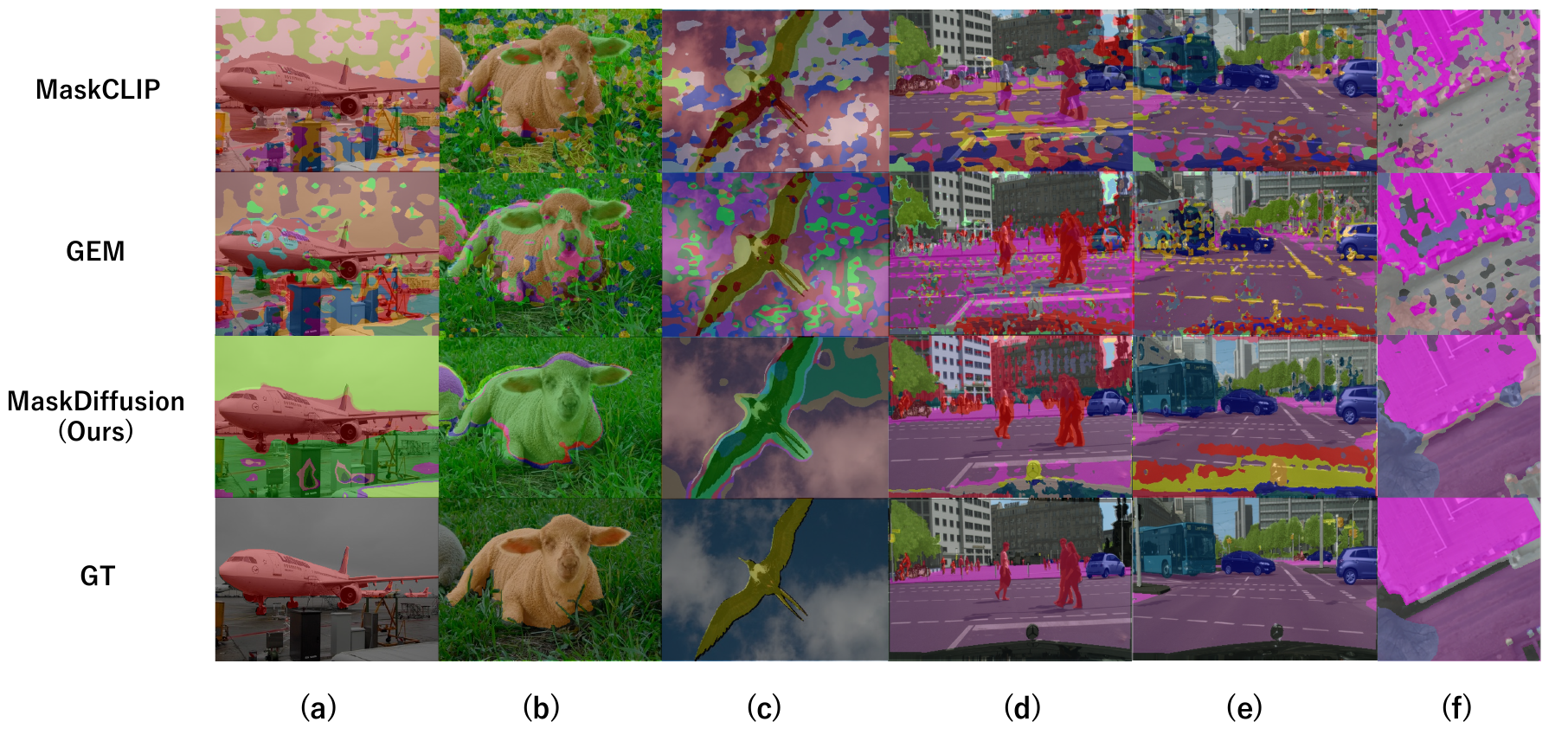}
    \centering
    \caption{\textbf{Qualitative results.} Images (a) to (c) depict scenes from PascalVOC~\cite{voc}, images (d) and (e) represent scenes from Cityscapes~\cite{cityscapes}, and image (f) shows scene from Potsdam~\cite{potsdam}.  We compare MaskCLIP~\cite{maskclip}, GEM~\cite{gem}, and our proposed MaskDiffusion. MaskCLIP~\cite{maskclip} and GEM~\cite{gem} provide fragmented and noisy segmentations, whereas MaskDiffusion exhibits a more cohesive and accurate segmentation for each object.}
    \label{fig:main}
    \end{minipage}
\end{figure*}

\subsection{Main Results}
\label{sec:main}
To validate the performance of MaskDiffusion, we conducted comprehensive comparative experiments with its closest counterpart, MaskCLIP~\cite{maskclip} and GEM~\cite{gem}.
For the evaluation, we used the mean Intersection over Union (mIoU) as the primary metric. We perform our experiments on the Potsdam~\cite{potsdam} dataset comprising of 6 classes, Cityscapes~\cite{cityscapes} with 19 classes, and PascalVOC~\cite{voc} with 20 classes, as well as COCO-Stuff~\cite{coco} consisting of 171 classes.
Note that we input the names of all the classes included in the dataset as prompts.

The quantitative results are presented in detail in Table~\ref{tab:main}.
Our method outperforms previous research across all datasets, particularly achieving a +10.5 mIoU on the Potsdam~\cite{potsdam} dataset, a land cover task dataset, demonstrating the robustness of the internal features for class estimation over CLIP-based approaches.

The qualitative results are shown in Figure~\ref{fig:main}.
Images (a) to (c) from PascalVOC~\cite{voc}, (d) and (e) from Cityscapes~\cite{cityscapes}, and (f) from Potsdam~\cite{potsdam} compare MaskCLIP~\cite{maskclip}, GEM~\cite{gem}, and our MaskDiffusion.
MaskCLIP~\cite{maskclip} and GEM~\cite{gem} result in a noisy and fragmented segmentation, whereas MaskDiffusion achieves more consistent segments that better correspond with the shape of the object and exhibit minimal noise.



To validate its performance, we conducted  experiments comparing our Unsupervised MaskDiffusion to its closest counterpart, DiffSeg~\cite{diffseg} under the same conditions of image unit segmentation as specified in their paper~\cite{diffseg}.
Following other unsupervised segmentation methods~\cite{picie, stego, hp, diffseg}, we quantify the performance of each method by employing the unsupervised mIoU, which matches unlabeled clusters with ground-truth labels using a Hungarian matching algorithm.
We test on Cityscapes~\cite{cityscapes} with 19 and 27 classes and COCO-Stuff~\cite{coco} with 27 classes. For a fair comparison, we input an empty string as a prompt.

The experimental results are shown in Table~\ref{tab:unsup}, which confirms that our proposed Unsupervised MaskDiffusion outperforms conventional segmentation methods using diffusion models and further underscores the effectiveness of internal features and their similarity map.

\begin{table*}[t]
  \caption{\textbf{Comparison of Unsupervised MaskDiffusion with state-of-the-art unsupervised segmentation methods.} We use the Cityscapes~\cite{cityscapes} and COCO-Stuff~\cite{coco} datasets for comparison and report the mIoU. The clear boost in performance by including the internal features underlines their semantic richness and effectiveness.}
  \label{tab:unsup}
  \centering
  \scalebox{1}[1]{
  \begin{tabular}{l | c | c | c c | c}
  \hline
    \multirow{2}{*}{Method} & \multirow{2}{*}{Backbone pretraining} & Training & \multicolumn{2}{c |}{Cityscapes~\cite{cityscapes}}& COCO-Stuff~\cite{coco}\\
    \cline{4-6}
     &  & free & 19class & 27class & 27class \\
     \hline
     IIC~\cite{iic} & - & - & - & 6.4 & 6.7 \\
     PiCIE~\cite{picie} & - & - & - & 12.3 & 13.8 \\
     STEGO~\cite{stego} & DINO~\cite{dino} & - & -  & 21.0 & 28.2\\
     HP~\cite{hp} & DINO~\cite{dino} & - & - & 18.4 & 24.6\\
     k-means & DINOv2~\cite{dinov2} & $\surd$ & 20.5 & 19.3 & 33.8 \\
     k-means & SD~\cite{sd} Internal feature & $\surd$ & 23.3 & 20.1 & 35.0 \\
     \hline
     DiffSeg~\cite{diffseg} & SD~\cite{sd} self-attention & $\surd$ & - & 21.2 & 43.6 \\
     MaskDiffusion (U) & SD~\cite{sd} Internal feature & $\surd$ & \textbf{28.5} & \textbf{25.3} & \textbf{58.4} \\
    \hline
  \end{tabular}
  }
\end{table*}
\begin{figure*}[t]
    \centering
    \begin{minipage}[b]{\linewidth}
    \includegraphics[width=\linewidth]{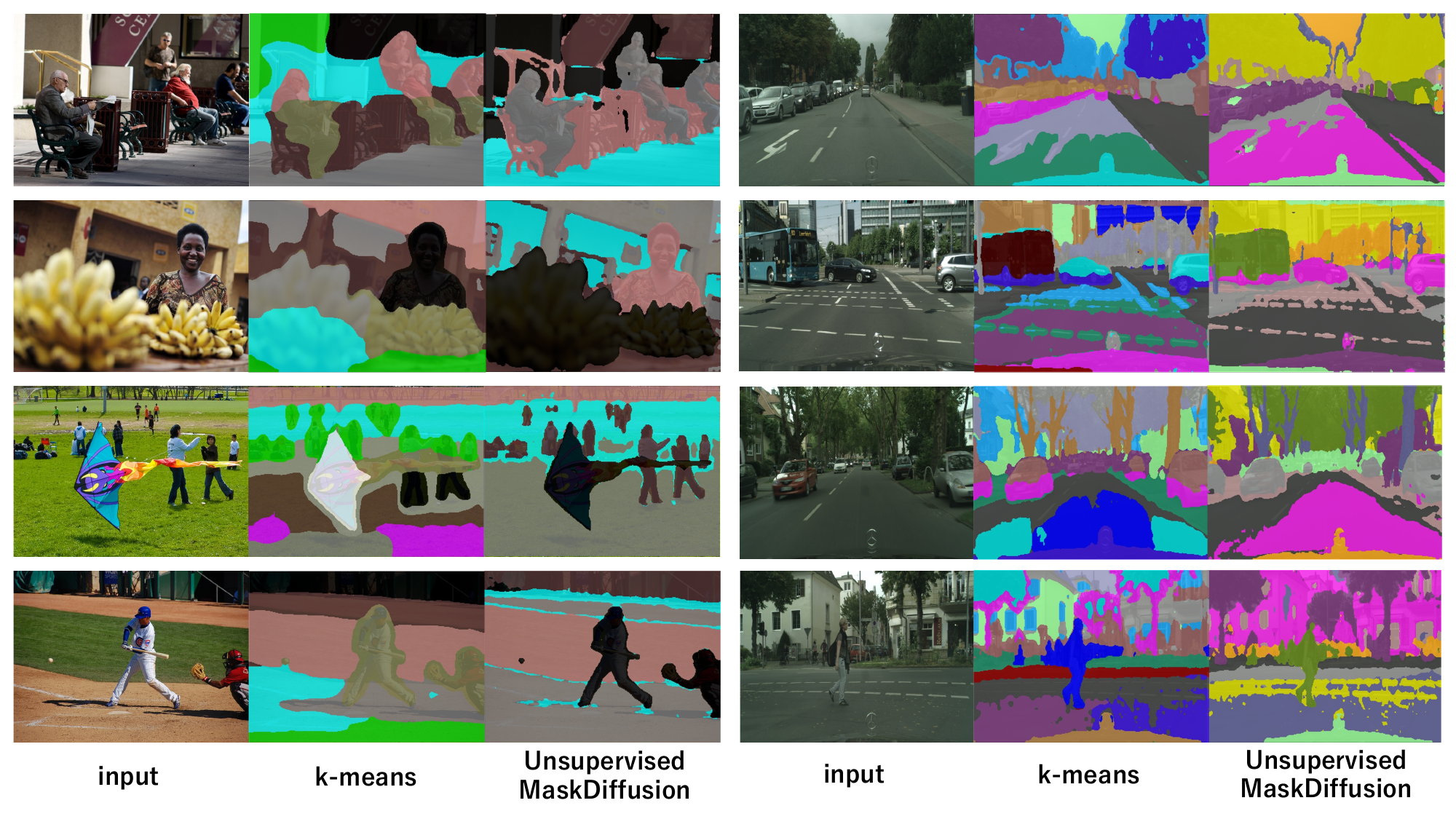}
    \caption{\textbf{Overview result of Unsupervised MaskDiffusion and k-means on internal features with COCO-Stuff~\cite{coco} and Cityscapes~\cite{cityscapes}.} It demonstrates that Unsupervised MaskDiffusion produces qualitatively cleaner results.}
    \label{fig:unsup_result}
    \end{minipage}
\end{figure*}

\subsection{Internal Feature}
\label{sec:internal}
In this section, we experimentally investigate the role of the internal features, which are represented by 1024-dimensional vectors for each pixel.
To prove that these internal features hold sufficient semantic information, we conduct k-means segmentation on a pixel-wise basis. To that end, we choose a specific number of k-means clusters for each dataset.

For comparison, we use common unsupervised segmentation methods~\cite{iic, picie, stego, hp} as well as k-means on DINOv2~\cite{dinov2}.
The experimental setup is the same as in the previous section.

The results of our experiments are presented in Table~\ref{tab:unsup}, highlighting the strong performance of the internal features across  diverse datasets. Particularly, the results demonstrate that applying k-means on the internal features significantly outperforms traditional unsupervised segmentation, for example by 6.8\% for the COCO-Stuff dataset. Moreover, the performance exceeds that of DINOv2 k-means across all datasets, showcasing the robust capabilities of the internal features without the need for specific training. 
Collectively, our findings establish the semantic nature of the internal features.

Figure~\ref{fig:unsup_result} shows the clustering results of Unsupervised MaskDiffusion and internal features via k-means. It demonstrates that Unsupervised MaskDiffusion can more cleanly classify identical classes into the same segment and different classes into separate segments, compared to k-means of internal feature.

\begin{figure*}[t]
    \begin{minipage}[b]{\linewidth}
    \includegraphics[width=0.9\linewidth]{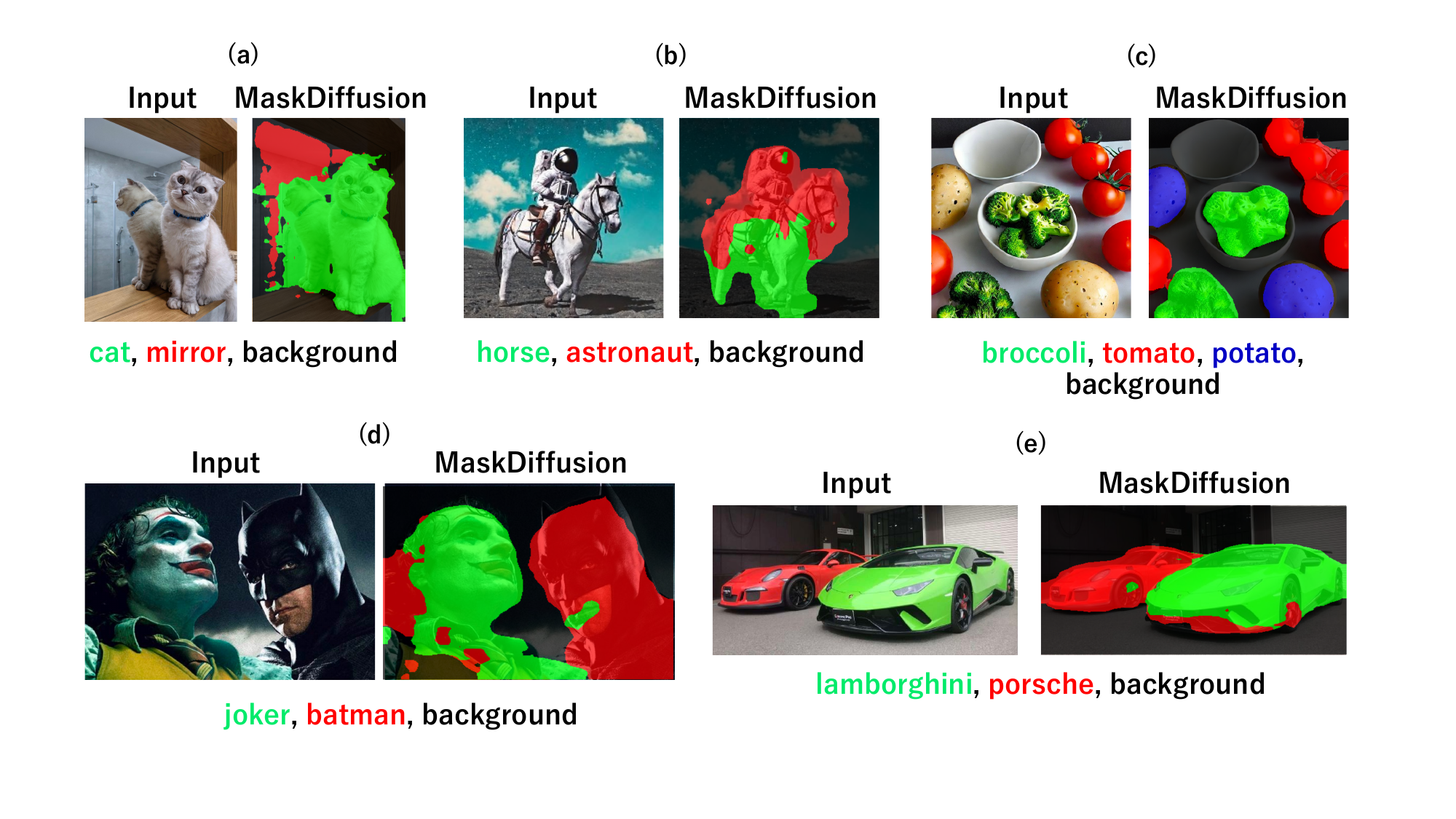}
    \centering
    \vspace{-1mm}
    \caption{\textbf{Open-vocabulary segmentation results.} In (a) we test on a general image, (b) and (c) show images generated by Stable Diffusion, and (d) and (e) are images featuring specific proper nouns. The corresponding prompt for each class is displayed below the respective image. The successful segmentation of challenging concepts, rare classes, and proper nouns highlights the effectiveness of MaskDiffusion in handling diverse segmentation tasks.}
    \label{fig:webimages}
    \end{minipage}
    \vspace{-1mm}
\end{figure*}

\subsection{Open Vocabulary Segmentation on Web-Crawled Images}
\label{sec:web}
In this section, we thoroughly assess the performance of MaskDiffusion using open vocabulary segmentation experiments on web-crawled images, where we test the model's ability to accurately segment various unseen classes, including specific and detailed categories such as 'joker' and 'porsche'.

The qualitative outcomes of the experiments are visually depicted in Figure~\ref{fig:webimages}. In this figure, (a) represents a general image, while (b) and (c) showcase images created by Stable Diffusion. Furthermore, (d) and (e) exhibit images containing proper nouns, with the respective class entered as a prompt displayed below each image.
Our results highlight the successful segmentation of challenging concepts such as 'mirror' in (a) and rare segmentation tasks such as 'astronaut' in (b). Additionally, the model demonstrates the capability to identify general classes, as depicted in (c), indicating that its segmentation performance improves with more general classes. Impressively, the segmentation of proper nouns is also achievable, as evidenced in results (d) and (e).
These findings serve as compelling evidence that MaskDiffusion exhibits a robust capability for accurately segmenting open vocabularies, including complex and specific categories, thus underscoring its versatility and effectiveness in handling diverse and intricate segmentation tasks.

\begin{figure}[t]
    \centering
    \includegraphics[width=0.9\linewidth]{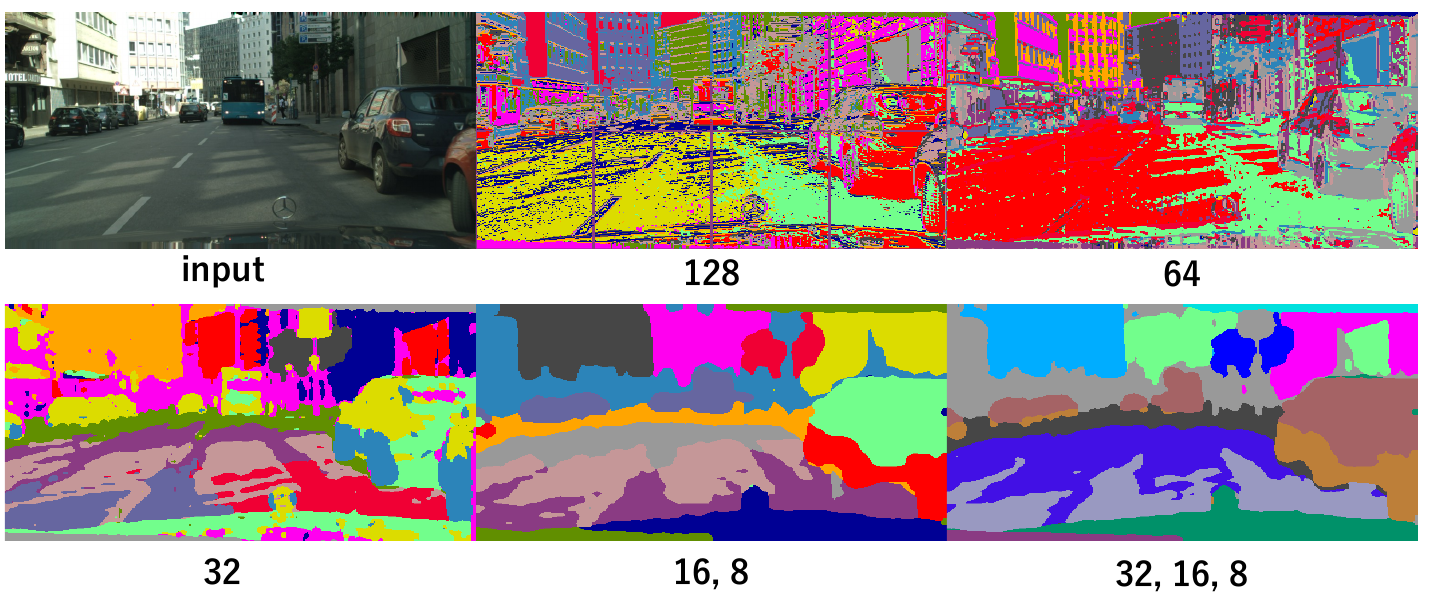}
    \caption{\textbf{Ablation study on internal feature size.} Qualitative results of internal features k-means show that earlier layers of the UNet (128, 64) do not provide an appropriate segmentation, whereas the inner layers (32, 16, 8) produce semantic meaningful results. 
    }
    \label{fig:ablation}
\end{figure}

\subsection{Ablation Study}
\label{sec:ablation}
In this section, we perform ablation studies on MaskDiffusion, examining how various Stable Diffusion~\cite{sd} time steps and UNet outputs for internal features affect performance, using the 19class Cityscapes~\cite{cityscapes} dataset.

Table~\ref{tab:ablation_t} presents the results of the ablation experiments comparing the effect of using MaskDiffusion with different time steps on the mIoU. The results reveal that employing a time step of $t=1$ yields the most favorable outcome. 
Typically, in Stable Diffusion~\cite{sd}, larger values of $t$ are employed when restoring images closer to Gaussian noise, requiring an attention map that broadly identifies the location of the object. Conversely, smaller values of $t$ are utilized to restore images closer to the original image, thereby producing an attention map that closely resembles the shape of the object in the real image.

Furthermore, we conduct ablation experiments on the cross-attention map and internal features, which involve examining the concatenated outputs of different layers. Considering the structure of UNet, the outer intermediate outputs are of larger size compared to the smaller inner intermediate outputs. 
Therefore, we compare various combinations at different positions, which are detailed in Table~\ref{tab:ablation_cmap}. Overall, combining the cross-attention maps with a resolution of $8 \times 8$ and $16 \times 16$ yields the best results.

Similarly, we examine the effect of combining different internal feature sizes by applying k-means and evaluating the resulting segmentation with the unsupervised mIoU. 
The quantitative results are summarized in Table~\ref{tab:ablation}, indicating that combining feature maps of size 8, 16, and 32 clearly outperforms other size combinations.
Figure~\ref{fig:ablation} presents the qualitative results of the ablation study.
Since earlier layers of the U-Net typically generate low-level features, the output is less suitable for segmentation, whereas inner outputs provide more semantic information. 

These findings highlight the critical role that both, the time step and the internal features, play in the segmentation performance of MaskDiffusion.

\begin{table}[t]
\centering
\begin{minipage}{0.45\linewidth}
\centering
\caption{\textbf{Ablation study on Diffusion time step.} We evaluate MaskDiffusion on the Cityscapes dataset~\cite{cityscapes} using mIoU.}
\label{tab:ablation_t}
\begin{tabular}{c | c c c c}
    \hline
      t & ~1~ & ~10~ & ~100~ & ~500~ \\
     \hline
     ~mIoU~ & \textbf{17.1} & 5.5 & 2.4 & 0.9\\
    \hline
\end{tabular}
\end{minipage}
\hfill
\begin{minipage}{0.45\linewidth}
\centering
\caption{\textbf{Ablation study on output size of the cross-attention map.} We evaluate MaskDiffusion on the Cityscapes dataset~\cite{cityscapes} using mIoU.}
\label{tab:ablation_cmap}
\vspace{-2mm}
\begin{tabular}{c c c | c }
    \hline
      \multicolumn{3}{c|}{Cross-attention map size}  & \multirow{2}{*}{mIoU} \\
      \cline{1-3}
      ~64~ & ~32 & 16,8 & \\
     \hline
     $\surd$ & $\surd$ & $\surd$ & 12.5 \\
     & $\surd$ & $\surd$ & 14.5\\
     & & $\surd$ & \textbf{17.1} \\
    \hline
\end{tabular}
\end{minipage}
\end{table}



Next, to explore the limitations of the internal features $\mathbf{f}$, we created representative $\mathbf{f}$ using images and ground truth on the training dataset.
In this experiment, $\mathbf{f}$ is first created from images in the training dataset, and then the $\mathbf{f}$ and ground truth are used to create a representative $\mathbf{f}$. Representative $\mathbf{f}$ is the mean of $\mathbf{f}$ for each class in ground truth.
In the same context, we also conduct an experiment using only the classes contained in the images as prompt input.
The evaluation results are shown in Table~\ref{tab:add_experiment}.

Our method achieves 30-60\% of the performance compared to when calculating representative $\mathbf{f}$ with ground truth, without training from the ground truth. Furthermore, it is observed that using dynamic prompts leads to improved performance in proportion to the number of classes per image.
Cityscapes~\cite{cityscapes}, COCO-Stuff~\cite{coco} contain about 10 objects per image, whereas VOC~\cite{voc} contains a few objects. mIoU increase is largely influenced by the number of objects in one image, especially with dynamic prompts.

\begin{table}[t]
  \caption{\textbf{Ablation study on U-Net layers to be adopted as internal features.} We perform k-means on various U-Net layer outputs and evaluate on the Cityscapes dataset~\cite{cityscapes}  using the unsupervised mIoU.}
  \vspace{-2mm}
  \label{tab:ablation}
  \centering
  \scalebox{1.0}[1.0]{
  \begin{tabular}{c c c c | c }
    \hline
      \multicolumn{4}{c|}{Internal feature size}  & Unsupervised \\
      \cline{1-4}
      ~128~ & ~64~ & ~32~ & 16,8 & mIoU\\
     \hline
     $\surd$ & & & & 13.8 \\
     & $\surd$ & & & 14.8 \\
     & & $\surd$ & & 17.2 \\
     & & & $\surd$ & 20.3 \\
     $\surd$ & $\surd$ & $\surd$ & $\surd$ & 14.8 \\
     & $\surd$ & $\surd$ & $\surd$ & 14.8 \\
     & & $\surd$ & $\surd$ & \textbf{23.3}\\
    \hline
  \end{tabular}
  }
\end{table}

\begin{table*}[t]
  \caption{\textbf{Additional experiment results of MaskDiffusion.} we conduct an experiment calculating representative $\mathbf{f}$ with ground truth and using only the classes contained in the images as prompt input (dynamic prompts).}
  \vspace{-2mm}
  \label{tab:add_experiment}
  \centering
  \scalebox{1.0}[1.0]{
  \begin{tabular}{l | c | c | c }
  \hline
    \multirow{2}{*}{Method} & Cityscapes~\cite{cityscapes}& PascalVOC~\cite{voc} & COCO-Stuff~\cite{coco}\\
    \cline{2-4}
     & 19class & 20class & 27class \\
     \hline
     MaskDiffusion & 17.1 & 29.9 & 13.0 \\
     MaskDiffusion w/ ground truth & 35.0 & 53.7 & 39.8 \\
     MaskDiffusion w/ dynamic prompts  & 21.6 & 87.2 & 40.5  \\
    \hline
  \end{tabular}
  }
\end{table*}

\section{Limitation}
While MaskDiffusion achieves remarkable results, especially compared to related approaches, our proposed method still comes with certain limitations. First, the cross-attention map has a low level of assigning internal features to each class. This can be seen in Figure~\ref{fig:main}(b), where the segmentation is clean but the class assignment is incorrect.
Second, our setting assumes that potential candidates for the classes appearing in the images are known beforehand. While we consider this to be outside of the scope of this paper, it would be possible to use MaskDiffusion in conjunction with models that are able to detect the presence of objects in images, e.g. CLIP~\cite{clip}, to solve this limitation.  

\section{Conclusion}
In this study, we proposed MaskDiffusion, a novel approach for semantic segmentation that leverages Stable Diffusion and internal features to achieve superior segmentation results. Through a series of comprehensive experiments and analyses, we have demonstrated the effectiveness and versatility of MaskDiffusion across various datasets and challenging segmentation tasks.
Our results indicate that MaskDiffusion exhibits robust performance in handling diverse categories, including general classes and fine-grained, proper noun-based segments.

Overall, our findings highlight the potential of MaskDiffusion as a powerful and effective tool in the field of semantic segmentation. By leveraging the strengths of Stable Diffusion and internal features, we have successfully demonstrated the model's capability to handle diverse datasets and open vocabularies, paving the way for future advancements and applications in this critical area of computer vision.

\clearpage  

%
%
\bibliographystyle{splncs04}
\bibliography{main}
\end{document}